%% file: AnonymousSubmission2027.tex
\documentclass[letterpaper]{article} 
\usepackage[preprint]{aaai2027} 
\usepackage[hyphens]{url}  
\usepackage{graphicx} 
\usepackage{natbib}  
\usepackage{caption} 
\usepackage{algorithm}
\usepackage{algorithmic}

\usepackage{newfloat}
\usepackage{listings}
\DeclareCaptionStyle{ruled}{labelfont=normalfont,labelsep=colon,strut=off} 
\floatstyle{ruled}
\newfloat{listing}{tb}{lst}{}
\floatname{listing}{Listing}

\usepackage{booktabs}

\usepackage{amsmath}
\usepackage{multirow}
\usepackage{amssymb}

\title{Dual-Path LLM Reasoning for Multimodal Few-Shot Knowledge Graph Completion}
\author{
    Jinlan Liu\textsuperscript{\rm 1},
    Zhiying Tu\textsuperscript{\rm 1,2,3}\corresponding,
    Yongchao Xing\textsuperscript{\rm 1},
    Yicheng Liu\textsuperscript{\rm 1},\\
    Bolin Zhang\textsuperscript{\rm 1,2,3},
    Dianbo Sui\textsuperscript{\rm 1,2},
    Dianhui Chu\textsuperscript{\rm 1,2},
    Hongliang Sun\textsuperscript{\rm 1,2,3}\corresponding
}

\affiliations{
    \textsuperscript{\rm 1}Harbin Institute of Technology (Weihai),
    Weihai 264209, China\\
    \textsuperscript{\rm 2}Shandong Provincial Key Laboratory of
    Digital Service Computing Technology and Systems,
    Weihai 264200, China\\
    \textsuperscript{\rm 3}Harbin Institute of Technology Qingdao Research Institute,
    Qingdao 266400, China\\
    26B903083@stu.hit.edu.cn,
    tzy\_hit@hit.edu.cn,
    \{22B903085, 25B903075\}@stu.hit.edu.cn
    \{bolin, suidianbo, chudh, sunhl\}@hit.edu.cn
}

\begin{document}

\maketitle

\begin{abstract}
Knowledge graph completion (KGC) aims to infer missing facts in knowledge graphs (KGs), thereby improving their completeness and supporting downstream intelligent applications.
However, emerging entities and relations in real-world deployments make inductive KGC difficult, especially under few-shot and zero-shot settings. Multimodal information and Large Language Model (LLM)-derived priors can enrich sparse relational contexts, but they may also introduce noisy or hallucinated evidence. 
To address these issues, we propose DuPLeR, a \textbf{Du}al-\textbf{P}ath \textbf{L}LM \textbf{R}easoning framework for multimodal few-shot KGC.
DuPLeR builds a calibrated relation graph by combining multimodal LLM-derived type priors with factual support structures, and performs dual-level structural reasoning over the refined relation topology. Moreover, a dual-pathway multimodal enhancement module regulates message passing with query-relevant multimodal signals and supplements entity representations after graph propagation. Experiments on eight inductive variants of two multimodal KG (MMKG) benchmarks show that DuPLeR achieves robust performance in data-scarce 
KGC scenarios.
\end{abstract}

\input{sections/introduction}

\input{sections/related_work}
\input{sections/preliminary}
\input{sections/method}

\input{sections/experiment}
\input{sections/conclusion}

\bibliography{aaai2027}

\input{sections/appendix}

\end{document}

%% file: sections/introduction.tex
\section{Introduction}

KGC, also known as link prediction, aims to infer missing facts in a KG, playing a critical role in maintaining the coverage and reliability of KGs \cite{10.1609/aaai.v39i14.33672}. 
Fact triples extracted automatically from massive 
logs, customer interactions, and so on are often incomplete, leaving potential relationships between  entities unlinked~\cite{HU2024110783}. This limits the effectiveness of KGs in enabling downstream tasks.
By performing KGC, missing links can be inferred and the KGs can be enriched, thereby enhancing their completeness and utility.

Most existing KGC methods operate in a transductive setting, where all entities and relations are observed during training and the goal is to predict missing links among them. 
However, the real world continuously witnesses the emergence of new entities that are incrementally added to the existing KG as new knowledge \cite{zhang2023neuralkgindpythonlibraryinductive}.
Traditional transductive KGC methods fail to update the knowledge representations of these newly added entities in a timely manner, requiring link prediction on unseen entities—known as inductive KGC \cite{zhang2024inductive}.
This capability is crucial for scalable intelligence and reducing annotation costs. 
Nevertheless, inductive KGC is fundamentally constrained by the scarcity of reliable contextual evidence for emerging entities, as emerging entities often provide only limited local neighborhoods and, in some cases, only a handful of support triples.

Multimodal information, such as textual descriptions and associated images of real-world items, can serve as auxiliary evidence to supplement the missing contextual information in sparse triples. By providing richer semantic signals beyond structural triples alone, multimodal cues enable models to better capture latent relationships between entities, thereby enhancing KGC performance.
Existing multimodal KGC methods have explored adaptive fusion mechanisms to selectively integrate heterogeneous modality-specific evidence \cite{MKGFORMER,LAFA,mmsn}, with some further incorporating multimodal neighborhood information to enrich entity or relation representations \cite{LAFA,mmsn}.
However, multimodal integration inevitably introduces cross-modal heterogeneity and noise, which can compromise the reliability of relational reasoning \cite{yang2025towards,kim2025kgmel,zhao2024contrast}.
This challenge is further amplified in inductive KGC settings. In scenarios involving emerging entities, the few-shot and data-scarce nature of the task means that noisy or misaligned modalities, especially visual or operational signals, may have a disproportionate impact on entity representations, distorting the representation space and ultimately degrading link prediction accuracy.

Given the scarcity of evidence that constrains inductive KGC, LLMs provide a promising way to enrich sparse relational contexts in low-resource settings, owing to their encoded world knowledge and strong generalization ability. 
Existing methods have explored the use of LLMs to extract relational context for KGC reasoning~\cite{prolink}, but they do not explicitly consider modality-specific reliability or granularity mismatch in MMKGs.
In MMKGs, the semantic priors derived from LLMs may also provide complementary high-level cues for bridging heterogeneous modalities and reducing noisy signals. However, directly combining LLM priors with multimodal information is still challenging. Different modalities describe relational contexts at different levels of granularity, and their reliability may vary significantly under limited supervision~\cite{mm-extract-part}. Therefore, the model should be able to selectively assess which signals are more trustworthy for a given prediction. 
This motivates a principled adaptive integration mechanism that can align LLM semantic priors with multimodal cues and improve inductive link prediction for emerging entities under few-shot and zero-shot conditions.

To address these challenges, we propose DuPLeR, a novel framework for inductive multimodal KGC in few-shot and zero-shot settings. First, we leverage LLMs for relation semantic enrichment to alleviate resource scarcity in few-shot scenarios. To mitigate model hallucination, we design a topology refinement module that utilizes factual knowledge to denoise and enrich the LLM-expanded information, thereby generating a calibrated relation graph for downstream reasoning. Subsequently, DuPLeR introduces a dual-pathway multimodal enhancement module, which leverages query-relevant and query-irrelevant multimodal information to inject multimodal cues during and after the message passing process, respectively, thereby selectively utilizing multimodal information to enhance entity representations. Finally, DuPLeR performs dual-level structural reasoning to simultaneously capture inter-relation patterns and entity-level structural evidence. 
We conduct extensive experiments on two public MMKG benchmarks, and the results demonstrate that DuPLeR effectively combines multimodal LLM-derived relational priors with multimodal signals to achieve more precise link prediction.
In summary, our paper makes the following contributions: 
\begin{itemize}

    \item We propose DuPLeR, which integrates multimodal LLM-derived relational priors with structural reasoning to mitigate representation distortion caused by sparse and noisy supervision, thereby improving inductive KGC.
    
    \item We design a dual-pathway multimodal enhancement mechanism that injects query-relevant multimodal evidence during message passing and incorporates query-independent multimodal semantics after propagation, enabling more robust multimodal fusion.

    \item Extensive experiments on eight variants of two MMKG benchmarks show that DuPLeR consistently improves inductive generalization and robustness.
\end{itemize}

%% file: sections/related_work.tex
\section{Related Work}

\subsection{Inductive KGC}

Inductive KGC aims to generalize link prediction to unseen entities and relations by learning transferable structural patterns. 
NBFNet captures relation-aware paths through graph message passing~\cite{nbfnet}, while InGram and ULTRA construct relation graphs to model transferable inter-relation dependencies~\cite{ingram,ultra}. 
Recent methods further introduce semantic knowledge to alleviate sparse structural evidence. 
ProLINK uses LLM-derived type priors to augment relation graphs for low-resource reasoning~\cite{prolink}, whereas SEMMA combines LLM-enriched relation semantics with structural relation graphs in fully inductive settings~\cite{arun2025semma}. 
However, these methods mainly focus on relation-level textual semantics. DuPLeR instead constructs a factually calibrated relation topology from multimodal type priors and further incorporates entity-level multimodal evidence.

\subsection{Multimodal KGC}

Multimodal KGC improves entity representations by combining structural, textual, and visual information. 
LAFA jointly models link-aware modality fusion and neighborhood aggregation~\cite{LAFA}, while MCKGC captures heterogeneous multimodal patterns in mixed-curvature spaces~\cite{MCKGC}. 
Recent approaches increasingly consider modality-specific reliability: NativE adopts relation-guided adaptive fusion to address modality imbalance~\cite{zhang2024native}, and MMSN combines multimodal neighborhood aggregation with meta-learning for few-shot KGC~\cite{mmsn}. 
In particular, multimodal LLMs are prone to being overly influenced by a single modality during content generation~\cite{zhuAnalyzingReasoningConsistency2026}.
Different from representation-level fusion, DuPLeR instead separates multimodal integration into query-conditioned regulation during structural message passing and holistic supplementation after graph propagation.

%% file: sections/preliminary.tex
\section{Problem Definition}

We study few-shot multimodal KGC under inductive settings. Let $\mathcal{G}=(\mathcal{E},\mathcal{R},\mathcal{S})$ denote an MMKG, where $\mathcal{S}\subseteq\mathcal{E}\times\mathcal{R}\times\mathcal{E}$ is the set of observed triples.
Each entity \( e\in\mathcal{E} \) is associated with multimodal side information, including a textual description \( t \) and a set of images \( \{ v_k \}_{k=1}^{K} \).
It is worth noting that under the settings of this paper, entities and relations appearing at test time are disjoint from those seen during training.
For a target relation \( r\in\mathcal{R} \), only a small support set \( \mathcal{S}_r=\{(h_i,r,t_i)\}_{i=1}^{N} \) with \( N\ll|\mathcal{S}| \) (or \( N=0 \)) is available.
Given a query request \( (q,r,?) \), the goal is to retrieve the appropriate tail entity by leveraging limited support triples.

%% file: sections/method.tex
\begin{figure*}[t]
  \centering
  \includegraphics[width=0.86\textwidth]{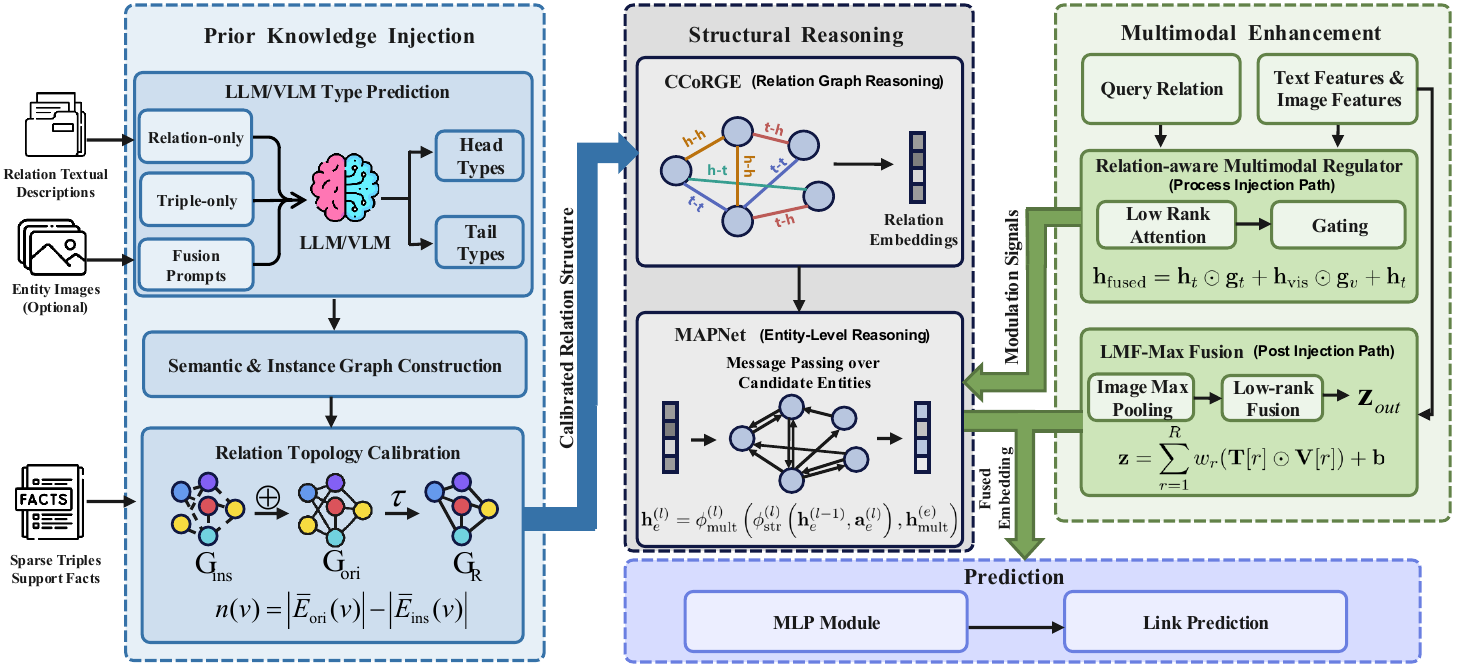}

  \caption{The overall framework of DuPLeR. It leverages multimodal LLM-derived relational priors for structural reasoning and employs a dual-path mechanism to inject multimodal signals, ultimately refining entity representations for link prediction.}
  
  \label{dupler_framework}
\end{figure*}

\section{Methodology}

In this section, we propose DuPLeR (Figure~\ref{dupler_framework}), a framework that leverages multimodal LLM-derived priors and entity-level multimodal features to improve KGC under scarce triplet resources.
Given a query relation, DuPLeR constructs a relation-aware context from sparse KGs and captures transferable relation--entity patterns through dual-level structural reasoning (Section~\ref{dual-level_structural_reasoning}).
This process is enhanced by multimodal LLM-derived relational priors, which provide semantic cues for relation topology construction and calibration (Section~\ref{llm-part}).
A dual-pathway multimodal module then injects entity-level multimodal information to refine representations and support robust entity matching (Section~\ref{multi-fusion}).
Finally, DuPLeR is pretrained on multiple MMKGs and generalizes to unseen graphs without retraining (Section~\ref{pretrain_reasoning}).

\subsection{Dual-level Structural Reasoning}
\label{dual-level_structural_reasoning}

In few-shot scenarios, each observed triple provides critical structural evidence. 
To effectively exploit such limited supervision, we perform dual-level structural reasoning over relations and entities. At the relation level, context-enhanced co-occurrence relation graph encoder (CCoRGE) constructs a high-order relation graph and performs graph propagation to capture transferable interaction patterns among relations. At the entity level, multimodal-aware prediction network (MAPNet) uses the resulting relation embeddings to guide message passing over the KG, enabling the model to identify relation-aware structural paths and score candidate entities for link prediction.
The two reasoning levels are complementary: the relation graph captures transferable interaction patterns that are less dependent on specific entity identities, whereas the entity graph grounds these patterns in query-specific factual paths.

\subsubsection{CCoRGE}
\label{relation_gnn}

The high-order relation graph is denoted as \(\mathcal{G}_R = (\mathcal{V}_R, \mathcal{E}_R)\), where the node set \(\mathcal{V}_R\) corresponds to the set of relations \(\mathcal{R}\) in \(\mathcal{G}\).
The edges in \(\mathcal{G}_R\) capture the topological co-occurrence regularities between relations. We define four distinct edge types, \(\mathcal{T} = \{h\text{-}h, h\text{-}t, t\text{-}h, t\text{-}t\}\), representing how two relations interact via shared entities. 
Formally, for \(a,b\in\{h,t\}\), a directed edge \((r_i,r_j,a\text{-}b)\in\mathcal{E}_R\) exists if the same entity appears at position \(a\) in a triple involving \(r_i\) and at position \(b\) in a triple involving \(r_j\).
For example, an edge $(r_i, r_j)$ of type \(\ h\text{-}t \) exists if there exists an entity $e \in \mathcal{E}$ such that:
\begin{equation}
    (e, r_i, x_i) \in \mathcal{S} \ \land\ (x_j, r_j, e) \in \mathcal{S}.
\end{equation}
During pre-training, \(\mathcal{G}_R\) is constructed from \(\mathcal{G}\). 
At reasoning time, it is augmented with semantic connections derived from the LLM-predicted head- and tail-entity types, as detailed in Section~\ref{llm-part}.
Finally, multi-layer GNN is employed to learn relation representations over \(\mathcal{G}_R\).

\subsubsection{MAPNet}

To ground transferable relation patterns in query-specific evidence while avoiding indiscriminate multimodal  fusion, MAPNet performs relation-aware propagation over the entity graph and incorporates the two multimodal  representations at complementary stages. 
During entity encoding, the initial representation of the source entity is obtained by fusing the CCoRGE representation of the query relation with the relation-aware multimodal representation of the source entity. All other entity states are initialized to zero. The CCoRGE representations of all relations are further projected into
layer-specific edge features for entity propagation.
Formally, at the \(l\)-th layer, MAPNet constructs a relation-aware message for each neighbor \(v \in \mathcal{N}(e)\) by concatenating its previous-layer representation \(\mathbf{h}_v^{(l-1)}\) with the corresponding relation embedding \(\mathbf{r}_{v,e}^{(l)}\), followed by a linear transformation parameterized by \(\mathbf{W}_M^{(l)}\). 
These messages are summed over all neighbors to obtain the neighborhood representation $\mathbf{a}_e^{(l)}$. The previous-layer state $\mathbf{h}_e^{(l-1)}$ and the aggregated neighborhood representation $\mathbf{a}_e^{(l)}$ are first structurally fused, after which the resulting representation is modulated by the multimodal context
$\mathbf{h}_{\mathrm{mult}}^{(e)}$:
\begin{equation}
\mathbf{h}_e^{(l)}
=
\phi_{\mathrm{mult}}^{(l)}
\left(
    \phi_{\mathrm{str}}^{(l)}
    \left(
        \mathbf{h}_e^{(l-1)},
        \mathbf{a}_e^{(l)}
    \right),
    \mathbf{h}_{\mathrm{mult}}^{(e)}
\right),
\label{entity_process_fusion}
\end{equation}
where $\phi_{\mathrm{str}}^{(l)}(\cdot)$ and $\phi_{\mathrm{mult}}^{(l)}(\cdot)$ denote the structural fusion and multimodal residual modulation at the $l$-th layer, respectively.
After \(L\) layers of structural propagation, the final score \(s(e)\) of entity \(e\) is computed by concatenating the hidden states from all GNN layers with the query-agnostic multimodal embedding \(\mathbf{z}_{\mathrm{out}}^{(e)}\):
\begin{equation}
\label{final_scoring}
s(e)
=
\operatorname{MLP}
\left(
\left(
\bigoplus\nolimits_{l=0}^{L}
\mathbf{h}_e^{(l)}
\right)
\parallel
\mathbf{z}_{\mathrm{out}}^{(e)}
\right),
\end{equation}
where \(\bigoplus\) denotes feature-wise concatenation across layers. 
The detailed constructions of \(\mathbf{h}_{\mathrm{mult}}^{(e)}\) and \(\mathbf{z}_{\mathrm{out}}^{(e)}\) are provided in Section~\ref{multi-fusion}.
Finally, we optimize the model using the standard binary cross-entropy (BCE) loss; its detailed formulation is provided in Appendix~\ref{train-obj}.

\subsection{Prior Knowledge Injection}
\label{llm-part}

\subsubsection{Multimodal LLM-based Type Predictor}
\begin{figure}[t]
  \centering
  \includegraphics[width=\linewidth]{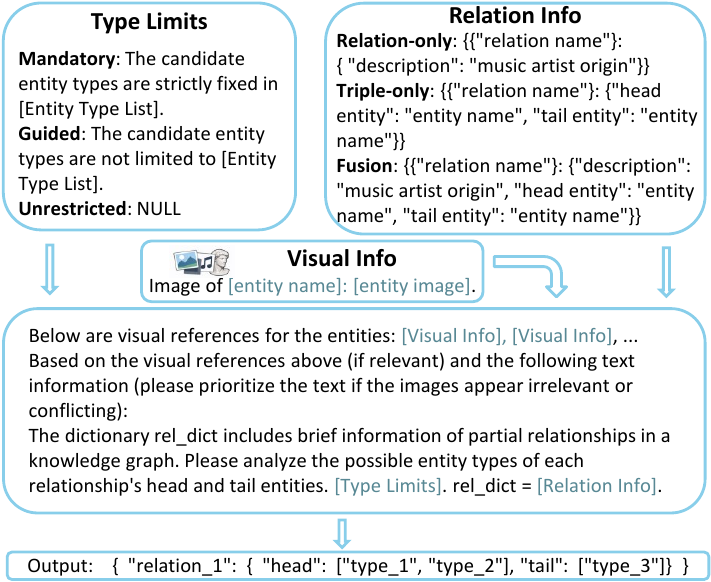}
  \caption{Multimodal LLM Prompts}
  \label{llm_prompt_tab}
\end{figure}

Relation modeling is pivotal to KGC. 
Under few-shot settings, the scarcity of relation-associated triples makes it difficult to learn reliable relation representations~\cite{FSLR}. 
This challenge becomes more pronounced in zero-shot scenarios, where conventional methods are often unable to infer meaningful representations for unseen relations without support triples~\cite{Generalizing_to_unseen_elements}.
To address this issue, we leverage carefully constructed prompts—augmented by sparse relational information—to infer the plausible types of head and tail entities, thereby enriching the relational graph and obtaining reliable relational representations.
The detailed prompts and output schema used for multimodal LLMs are illustrated in Figure~\ref{llm_prompt_tab}.
Specifically, entity images are treated as optional evidence rather than uniformly reliable cues. Since images may contain abundant details that are irrelevant to the target relation, the prompt instructs the multimodal LLM to use visual information only when it is relevant to the relation and consistent with the accompanying textual context, and to otherwise ignore it, thereby reducing visual noise. For pure-text LLMs, the visual inputs are omitted while the remaining prompt structure is retained. To regularize type inference, we constrain the outputs using a generic entity type list under three progressively relaxed settings: Mandatory, Guided, and Unrestricted. For each relation, the model returns a list of plausible types for the head entities and another list for the tail entities. We further provide relational context in three formats---Relation-only, Triple-only, and Fusion---to examine the trade-off between type inference quality and prompting overhead. 
The predicted head- and tail-type sets of each relation are used to construct \(\mathcal{G}_{\text{sem}}\) based on type-set overlap, following the positional connectivity scheme of CCoRGE (Section~\ref{relation_gnn}).
Overall, this prompting strategy aims to enrich sparse relational contexts and provide reliable semantic priors for downstream inductive KGC.

\subsubsection{Relational Topology Refinement}
To mitigate hallucinations in multimodal LLMs, we introduce a factual-triple-based refinement step that filters potentially spurious outputs and enriches them with reliable information.
We begin by constructing a structural relation graph \(\mathcal{G}_{\text{ins}}\) for the target relation \(r_q\) based on the support triples and the background KG, following Section~\ref{relation_gnn}. 
We then merge \(\mathcal{G}_{\text{sem}}\) and \(\mathcal{G}_{\text{ins}}\)
to obtain the preliminary relation graph \(\mathcal{G}_{\text{ori}}\).
To reduce noisy relation neighbors introduced during graph augmentation, we apply an indirect noise filtering strategy to the first-order neighbors of \(r_q\) in \(\mathcal{G}_{\text{ori}}\). 
For each neighbor \(v \in \mathcal{N}_{\text{ori}}(r_q)\), we measure its noise degree by comparing its non-target connections in the merged graph and the structural graph:
\begin{equation}
    n(v) = 
    \left| \bar{E}_{\text{ori}}(v) \right|
    -
    \left| \bar{E}_{\text{ins}}(v) \right|,
\end{equation}
where \(\bar{E}_{\text{ori}}(v)\) and \(\bar{E}_{\text{ins}}(v)\) denote the sets of edges incident to \(v\) in \(\mathcal{G}_{\text{ori}}\) and \(\mathcal{G}_{\text{ins}}\), respectively, excluding the edge connecting \(v\) to the target relation \(r_q\). 
Given a predefined noise threshold \(\tau\), the edges connecting \(r_q\) to neighbors with \(n(v) \geq \tau\) are collected into the noisy-edge set:
\begin{equation}
\mathcal{E}_{\mathrm{noise}}
=
\left\{
(r_q,v) \in \mathcal{E}_{\mathrm{ori}}
\mid
v \in \mathcal{N}_{\mathrm{ori}}(r_q),
\;
n(v) \geq \tau
\right\}.
\end{equation}
The final denoised relation graph \( \mathcal{G}_{R} \) is obtained by removing these edges from the merged relation graph:
\begin{equation}
\mathcal{G}_{R}
=
\left(
\mathcal{V}_{\mathrm{ori}},
\mathcal{E}_{\mathrm{ori}}
\setminus
\mathcal{E}_{\mathrm{noise}}
\right).
\end{equation}
A larger \(\tau\) retains more auxiliary semantic connections, whereas a smaller \(\tau\) results in more aggressive noise suppression.

\subsection{Dual-pathway Multimodal Enhancement}
\label{multi-fusion}

While relation graphs inferred by multimodal LLMs offer rich relational context, 
they remain unreliable as standalone evidence and must be anchored by structural message passing~\cite{llm_disable}.
To complement this LLM-derived relational context with grounded entity-level semantics, multimodal information is essential; however, 
indiscriminate fusion may introduce substantial query-irrelevant noise and distort structural message passing~\cite{zhangMultimodalFusionLowquality2024,mm-extract-part}.
Injecting query-specific multimodal information during message passing can suppress modality cues irrelevant to the current relation, but such relation-conditioned filtering may discard global semantics useful for entity identification. Conversely, query-agnostic fusion preserves comprehensive entity-level semantics, yet directly introducing these unfiltered features into message passing may interfere with relation-specific structural paths.
Neither pathway alone can simultaneously maintain relation-aware structural reasoning and preserve complete entity-level semantics.
Therefore, DuPLeR employs two complementary pathways: Query-Specific Regulation provides controlled multimodal guidance during structural propagation, while Query-Agnostic LMF-Max compensates for the semantic loss induced by relation-conditioned filtering by restoring comprehensive entity-level semantics after propagation.

\subsubsection{Query-Specific Regulation}
\label{message_passing_regulator}
To instantiate the relation-aware regulation, we dynamically fuse text and visual features conditioned on the query relation. After projecting normalized features into a shared space, we obtain a unified visual representation $\mathbf{h}_{\text{vis}}$ via relation-guided low-rank attention. 
The $k$-th image attention weight is:
\begin{equation}
\alpha_k=
\operatorname{softmax}_{k}
\left(
(\mathbf{W}_q\mathbf{q})^\top
(\mathbf{W}_v\mathbf{f}_{v_k})
\right),
\end{equation}
where $\mathbf{W}_q$ and $\mathbf{W}_v$ are low-rank projection matrices.
To precisely control modality contributions, we deploy modality-specific low-rank gates, $\mathbf{g}_t$ and $\mathbf{g}_v$, which are parameterized by the query relation:
\begin{equation}
    \mathbf{h}_{\text{fused}} = \mathbf{h}_t \odot \mathbf{g}_t + \mathbf{h}_{\text{vis}} \odot \mathbf{g}_v + \mathbf{h}_t .
\end{equation}
Unlike standard gating mechanisms, our relation-aware gating coupled with a residual text connection is specifically designed to alleviate modality imbalance. We then apply layer normalization and dropout to $\mathbf{h}_{\text{fused}}$ to obtain the relation-guided multimodal embedding $\mathbf{h}_{\text{mult}}$.
Rather than directly appending multimodal features, which may dilute graph-topological information, we use \(\mathbf{h}_{\text{mult}}\) as a soft regulator to modulate the structurally updated entity representation at each layer, as formulated in Equation~\eqref{entity_process_fusion}.
This customized modulation ensures that structural reasoning remains dominant while being smoothly guided by query-relevant multimodal cues. A formal proof is provided in the supplementary material.

\subsubsection{Query-Agnostic LMF-Max}

To compensate for the aforementioned information bottleneck, our post-fusion pathway captures salient visual evidence while excluding invalid image inputs. Specifically, invalid image positions indicated by the validity mask are filled with a large negative value before dimension-wise max pooling, while entities with no valid images are assigned a zero visual representation, yielding $\tilde{\mathbf{f}}_v$. We then project the textual feature and $\tilde{\mathbf{f}}_v$ into a low-rank space, obtaining $\mathbf{T}, \mathbf{V} \in \mathbb{R}^{R \times d}$.
Unlike vanilla LMF, which treats all low-rank components equally during aggregation, LMF-Max regards each rank as a distinct semantic subspace and assigns it a learnable weight $\mathbf{w} \in \mathbb{R}^{R}$:
\begin{equation}
    \mathbf{z}
    =
    \sum\nolimits_{r=1}^{R}
    w_r
    \left(
        \mathbf{T}[r] \odot \mathbf{V}[r]
    \right)
    + \mathbf{b}.
\end{equation}
Masked max pooling preserves the most salient valid visual signals, while rank-wise weighting selectively emphasizes informative cross-modal interactions. This pathway remains query-agnostic, providing complementary entity-level semantics that are incorporated into the final prediction as formulated in Equation~\eqref{final_scoring}.

\subsection{Pre-training and Inference}
\label{pretrain_reasoning}

During pre-training, DuPLeR is jointly optimized across multiple multimodal knowledge graphs, with all trainable components fully shared across graphs. In addition, to balance the varying difficulty levels across graphs, DuPLeR adopts a loss-guided dynamic dataset weighting strategy, as detailed in Appendix~\ref{sec:dynamic_sampling}.

During inference, DuPLeR derives representations of unseen relations based on the outputs of multimodal LLMs. These relation representations are subsequently injected into the knowledge graph to obtain representations of unseen entities. Consequently, DuPLeR requires no retraining on the target graph, thereby enabling more precise and context-aware link prediction under data-scarce conditions.

%% file: sections/experiment.tex
\section{Experiment}






\begin{table*}[t]
\centering
\begin{tabular}{llcccccccc}
\hline
Dataset & Model 
& \multicolumn{2}{c}{v1} 
& \multicolumn{2}{c}{v2} 
& \multicolumn{2}{c}{v3} 
& \multicolumn{2}{c}{v4} \\
\cline{3-4} \cline{5-6} \cline{7-8} \cline{9-10}
 &  & 1s & 3s & 1s & 3s & 1s & 3s & 1s & 3s \\
\hline

\multirow{12}{*}{FB15K-IMG-R}
& DistMult   & 0.07 & 0.04 & 1.00 & 1.80 & 0.71 & 1.77 & 2.00 & 3.38 \\
& RotatE     & 2.19 & 6.00 & 1.69 & 2.23 & 1.15 & 2.50 & 2.74 & 2.96 \\
& InGram     & 14.72 & 17.48 & 6.11 & 7.32 & 17.86 & 19.33 & 13.47 & 14.23 \\
& ULTRA(3g)  & 27.00 & 27.10 & 10.10 & 11.13 & 12.70 & 10.30 & 6.63 & 6.70 \\
& MMSN       & 14.70 & 22.20 & 45.00 & 39.90 & 21.90 & 29.33 & 21.70 & 27.90 \\
& ProLINK -- Llama2-7B   
& 53.80 & 59.00 & \underline{51.50} & \underline{53.30} & 50.00 & 51.40 & 41.40 & 42.20 \\
& ProLINK -- Qwen3-8B    
& \underline{61.40} & \underline{62.83} & 49.87 & 52.23 & 50.80 & 51.77 & 41.70 & 42.23 \\
& ProLINK -- DeepSeek-V3.2 & 61.20 & 62.27 & 50.13 & 51.80 & \underline{52.70} & \underline{52.30} & \underline{41.87} & \underline{43.10} \\
& \textbf{DuPLeR -- Llama2-7B}      
& 69.73 & 72.20 & \textbf{58.43} & \textbf{60.10} & 64.57 & \textbf{66.17} & 47.60 & \textbf{48.47} \\
& \textbf{DuPLeR -- Qwen3-8B}       
& 71.63 & \textbf{73.20} & 57.97 & 59.93 & 64.90 & 65.90 & 47.60 & 48.40 \\
& \textbf{DuPLeR -- Qwen3-VL-8B}    
& 71.67 & 73.13 & 58.40 & 60.03 & \textbf{65.20} & 66.03 & \textbf{47.97} & 48.37 \\
& \textbf{DuPLeR -- DeepSeek-V3.2}  
& \textbf{71.77} & 73.00 & 58.23 & 60.00 & 65.17 & 66.07 & 47.50 & \textbf{48.47} \\
\hline

\multirow{12}{*}{OpenBG-IMG-R}
& DistMult   & 0.00 & 0.02 & 0.73 & 1.06 & 0.98 & 1.92 & 0.88 & 1.77 \\
& RotatE     & 0.00 & 0.00 & 0.02 & 0.68 & 0.01 & 0.94 & 0.02 & 0.84 \\
& InGram     & 5.37 & 6.07 & 2.22 & 3.23 & 8.09 & 9.58 & 0.45 & 1.42 \\
& ULTRA(3g)  & 6.37 & 6.70 & 3.53 & 3.13 & 1.70 & 1.17 & 2.13 & 2.43 \\
& MMSN       & 11.17 & 9.13 & 5.10 & 1.63 & \underline{16.90} & \underline{20.50} & \underline{22.60} & 14.60 \\
& ProLINK -- Llama2-7B   
& 20.50 & 23.30 & 14.40 & 16.43 & 14.00 & 18.50 & 18.90 & \underline{24.00} \\
& ProLINK -- Qwen3-8B    
& 22.37 & 25.67 & 10.20 & 14.37 & 11.47 & 17.50 & 17.03 & 23.40 \\
& ProLINK -- DeepSeek-V3.2 & \underline{22.60} & \underline{27.03} & \underline{15.27} & \underline{21.83} & 12.97 & 18.03 & 17.63 & 23.43 \\
& \textbf{DuPLeR -- Llama2-7B}      
& 17.33 & 20.97 & 18.07 & \textbf{23.30} & 19.13 & \textbf{28.17} & \textbf{23.07} & 29.70 \\
& \textbf{DuPLeR -- Qwen3-8B}       
& \textbf{25.73} & \textbf{30.13} & 17.03 & 22.90 & \textbf{21.10} & 27.53 & 22.93 & 29.37 \\
& \textbf{DuPLeR -- Qwen3-VL-8B}    
& 17.40 & 21.03 & \textbf{18.17} & \textbf{23.30} & 19.07 & \textbf{28.17} & 22.83 & 29.70 \\
& \textbf{DuPLeR -- DeepSeek-V3.2}  
& 17.53 & 21.37 & 17.53 & 23.00 & 20.27 & 27.97 & \textbf{23.07} & \textbf{30.00} \\
\hline
\end{tabular}%
\caption{Few-shot Hits@10 results (\%) on FB15K-IMG-R and OpenBG-IMG-R, where 1s and 3s denote the 1-shot and 3-shot settings. Bold and underline indicate the best overall and best baseline results, respectively. }
\label{main_experiment_results}
\end{table*}

\subsection{Experiment Settings}
\subsubsection{Datasets}

To evaluate DuPLeR, we use five MMKG datasets: FB15K-IMG-R, DB15K-EMB-R, and YAGO15K-EMB-R derived from MMKB~\cite{mmkb-numerical,mmkb-visual}; OpenBG-IMG-R from the OpenBG benchmarks~\cite{ICDE2023_OpenBG}; and CoDEx-IMG-R, which we construct from the publicly available CoDEx dataset~\cite{safavi-koutra-2020-codex}. 
Here, the suffix ``EMB'' denotes datasets in which the original images are represented by pre-extracted visual embeddings, whereas ``IMG'' denotes datasets that provide the raw image files.

DB15K-EMB-R, YAGO15K-EMB-R, and CoDEx-IMG-R are used for pre-training with standard transductive splits. 
FB15K-IMG-R and OpenBG-IMG-R are used for few-shot link prediction. 
Following InGram~\cite{ingram}, we create four variants for each dataset according to the proportions of triples involving new relations in the inference graph, with ratios of 0.25, 0.50, 0.75, and 1.00. 
For each variant, three random seeds are used to generate 0-shot, 1-shot, and 3-shot settings, resulting in 72 few-shot datasets for evaluation.

\subsubsection{Baselines}

To evaluate DuPLeR, we compare it with three groups of baselines:
(1) \textbf{Traditional KG Embedding methods}: DistMult~\cite{distmult}, which uses diagonal relation matrices for bilinear scoring, and RotatE~\cite{sun2018rotate}, which represents relations as rotations in complex space.
(2) \textbf{Unimodal models}: InGram~\cite{ingram}, which learns relation and entity representations by aggregating neighborhood information over relation-weighted KG structures, and ULTRA~\cite{ultra}, which builds relation graphs from triples to capture inter-relation dependencies, and ProLINK~\cite{prolink}, which extends ULTRA with LLM-derived priors for few-shot reasoning.
(3) \textbf{Multimodal models}: MMSN~\cite{mmsn}, which combines multimodal neighbor aggregation with meta-learning-based cross-modal interaction modeling.

\subsubsection{Experimental Settings}
In the comparative experiments, we followed the original settings of the baseline models and adapted them for the few-shot dataset setup. DuPLeR is implemented with a three-layer DistMult-based GNN for relation encoding and a six-layer RotatE-based GNN for entity encoding. The low-rank fusion rank, bottleneck dimension, and embedding dimension are set to 2, 64, and 64, respectively. 
The model is trained with AdamW using an initial learning rate of 0.003 and a cosine annealing scheduler.
For evaluation, we use three LLMs, Llama2-7B~\cite{Touvron2023Llama2O}, Qwen3-8B~\cite{qwen3technicalreport}, and DeepSeek-V3.2~\cite{deepseek-aiDeepSeekV32PushingFrontier2025}, along with the VLM Qwen3-VL-8B-Instruct~\cite{qwen3technicalreport}, abbreviated as Qwen3-VL-8B in subsequent tables.
We use Hits@10 as the evaluation metric, which measures the proportion of test queries whose ground-truth tail entity appears among the top 10 predicted entities.

\subsection{Main Results}

We conducted experiments under the aforementioned settings, and the results are presented in Table~\ref{main_experiment_results}.
It should be noted that, for traditional KG Embedding models, we employed the OpenKE toolkit~\cite{han2018openke}. All results reported in the table are averaged over three runs with different random seeds on the few-shot datasets.

The following conclusions can be drawn from the table: 
(1) Superiority of LLM Integration: ProLINK and DuPLeR significantly outperform traditional baselines across most settings. For instance, on FB15K-IMG-R v1, DuPLeR achieves 71.77\% compared to ULTRA’s 27\%, validating that LLM-injected prior knowledge effectively enriches sparse context for KGC.
(2) Effectiveness of Dual-Path Enhancement: DuPLeR generally surpasses ProLINK, demonstrating that the dual-path multimodal enhancement module successfully leverages entity-level multimodal information to complement the relational context provided by LLMs.
(3) Model robustness and backbone diversity: Peak performance varies across different LLM backbones (e.g., DeepSeek-V3.2 leads in only specific cases). This diversity reflects differences in pre-training corpora and highlights DuPLeR’s robustness; it maintains high Hits@10 performance via multimodal cues even when LLM-derived relational context is insufficient, thereby validating the effectiveness of the dual-path multimodal enhancement module.

\subsection{Zero-Shot Performance}

The motivation behind incorporating multimodal LLMs is to alleviate information scarcity and provide high-quality relational representations—an effect that becomes particularly pronounced in a zero-shot setting. To demonstrate the generalizability of these representations across broad world knowledge,
we evaluated the zero-shot performance on FB15K-IMG-R. The results comparing several baseline models alongside DuPLeR are presented in Table~\ref{main_experiment_0}.
It is worth noting that, under this specific setting, the prompts provided to all models consisted solely of textual information containing the relation descriptions.

\begin{table}[thb]
\centering
\begin{tabular}{lcccc}
\hline
 & \multicolumn{4}{c}{\textbf{FB15K-IMG-R}} \\
\cline{2-5}
Model & v1 & v2 & v3 & v4 \\
\hline
InGram         & 13.56 & 3.41  & 16.02 & 12.90 \\
ULTRA(3g)      & 25.90 & 13.90 & 21.10 & 9.60  \\
MMSN           & 0.30  & \underline{38.10} & 11.60 & 24.80 \\
\hline
ProLINK &&&& \\
-- Llama2-7B   & 27.90 & 12.70 & 10.50 & 7.80  \\
-- Qwen3-8B    & \underline{58.50} & 33.10 & 37.00 & 25.90 \\
-- DeepSeek-V3.2 & 57.90 & 36.20 & \underline{43.30} & \underline{29.70} \\
\hline
\textbf{DuPLeR} &&&& \\
\textbf{-- Llama2-7B}   & 61.40 & 35.60 & 37.50 & 25.30 \\
\textbf{-- Qwen3-8B}    & \textbf{64.30} & 39.90 & 44.10 & 32.10 \\
\textbf{-- Qwen3-VL-8B} & 62.60 & 39.80 & 52.70 & 30.10 \\
\textbf{-- DeepSeek-V3.2} & 64.00 & \textbf{42.70} & \textbf{56.40} & \textbf{34.70} \\
\hline
\end{tabular}
\caption{0-shot evaluation results on FB15K-IMG-R (\%). Bold numbers indicate the best overall performance; underlined numbers indicate the best among baseline models.}
\label{main_experiment_0}
\end{table}

The zero-shot experimental results corroborate the conclusions of the main experiments—namely, that incorporating LLM prior knowledge effectively mitigates the problem of information sparsity.
Notably, DuPLeR shows a clear improvement over ProLINK, which also utilizes LLMs, with the enhancement being particularly pronounced for the relatively weaker Llama2-7B model. Furthermore, in the v1 version, the performance achieved using the compact Qwen3-8B model surpassed that of the full-scale DeepSeek-V3.2 model. This suggests that DuPLeR's dual-path multimodal information injection module effectively complements the relational context enriched by LLMs, thereby further mitigating the issue of information scarcity within a zero-shot setting and enabling more precise predictions.

\subsection{Ablation Study}

\begin{table}[tbh]
\centering
\begin{tabular}{lccc}
\hline
Model & \multicolumn{3}{c}{\textbf{OpenBG-IMG-R V1}} \\
\cline{2-4}
 & 0-shot & 1-shot & 3-shot \\
\hline
\textbf{DuPLeR}       & \textbf{3.90} & \textbf{17.40} & \textbf{21.03} \\
w/o - multimodal LLM  & 2.80    & 16.23    & 20.10          \\
w/o - Message Passing & 0.80    & 12.80    & 17.50          \\
w/o - LMF-Max             & 2.70    & 12.80    & 17.97          \\
\hline
\end{tabular}
\caption{Ablation evaluation results on the OpenBG-IMG-R V1 dataset (\%) with Qwen3-VL-8B.}
\label{ablation_result}
\end{table}

Constrained by page limits, we conduct our ablation studies exclusively on OpenBG-IMG-R V1 using Hits@10 under 0-shot, 1-shot, and 3-shot settings.
Three DuPLeR variants are evaluated: w/o multimodal LLM, which removes multimodal LLM prompting during evaluation; w/o Message Passing, which disables multimodal process injection during pre-training; and w/o LMF-Max, which disables multimodal post-injection during pre-training. These variants examine the contribution of large-model relational knowledge and multimodal injection strategies.
As shown in Table~\ref{ablation_result}, all modules contribute to DuPLeR. Removing the multimodal LLM prior causes a $\sim28.2\%$ relative performance drop in zero-shot setting, highlighting its vital role in enriching relational context without reference examples. This drop diminishes in few-shot scenarios, where explicit empirical demonstrations partially replace the LLM's semantic prior, leading to a diminishing marginal effect. Notably, this degradation is less severe than ablating the Dual-pathway Multimodal Enhancement module. This suggests robust multimodal entity representations can partially offset limited relational context, underscoring the dual-path injection's importance in capturing multimodal signals.

\subsection{Hyperparameter Analysis}

\begin{figure}[tb]
  \centering
  \includegraphics[width=\linewidth]{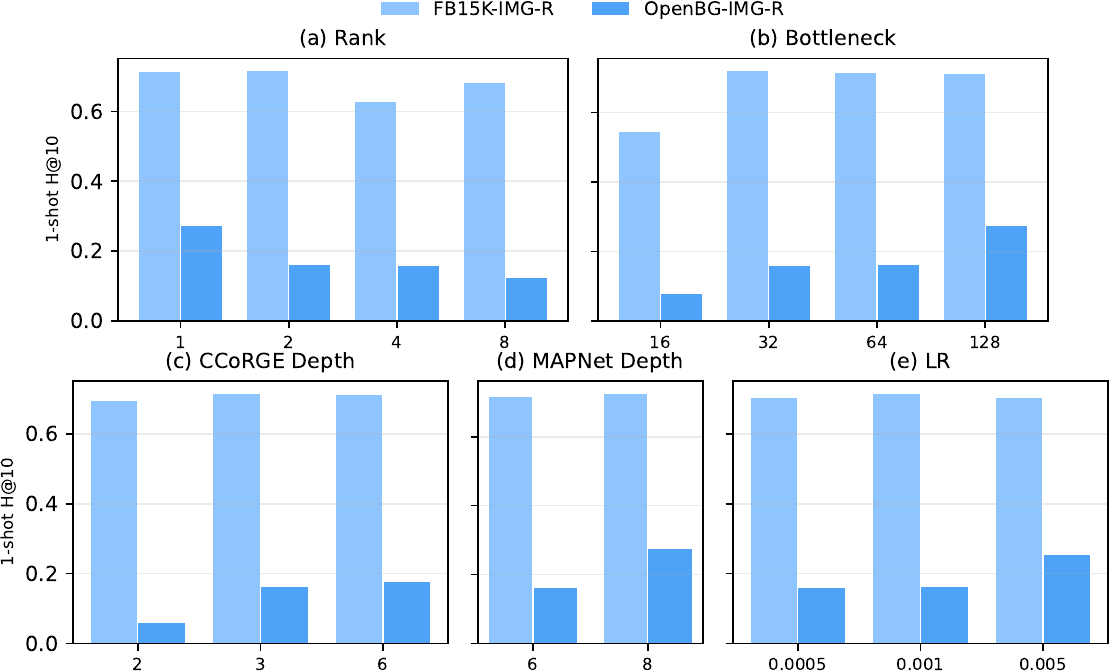}
  \caption{Hyperparameter sensitivity analysis. 
  We analyze 1-shot Hits@10 performance on FB15K-IMG-R V1 and OpenBG-IMG-R V1 while varying one hyperparameter at a time (low-rank rank, bottleneck dimension, CCoRGE depth, MAPNet depth, learning rate).
  }
  \label{hyper_fig}
\end{figure}

To systematically evaluate the impact of hyperparameter variations on DuPLeR, we analyze four architecture-related hyperparameters—bottleneck dimension, low-rank decomposition rank, CCoRGE depth, and MAPNet depth—as well as the training hyperparameter learning rate. 
In our analysis, we select the FB15K-IMG-R V1 and OpenBG-IMG-R V1 datasets and report performance under the 1-shot setting using the Hits@10 metric, as 1-shot is more sensitive compared to 0-shot and 3-shot. The results are shown in Figure~\ref{hyper_fig}.

As shown in the figure, adjusting the learning rate has minimal impact on FB15K but substantially improves OpenBG performance, indicating that a larger learning rate can enhance generalization. Increasing the bottleneck dimension also improves performance, while too small a dimension (e.g., 16) excessively compresses multimodal semantics, weakening relation-aware gating and structural reasoning. The rank of low-rank interaction affects few-shot generalization: lower rank constrains cross-modal interactions and introduces stronger structural bias, whereas excessively large rank may cause overfitting. CCoRGE depth is critical for modeling relational context; shallow networks fail to capture sufficient context, while overly deep networks may overfit on simpler datasets. 
MAPNet depth has a smaller but consistently positive effect, expanding the structural receptive field and enabling richer path reasoning for more precise link predictions.

%% file: sections/conclusion.tex
\section{Conclusion}

This paper proposes DuPLeR, a novel framework for few-shot MMKG completion. DuPLeR combines multimodal LLM-derived relational priors, topology-refined structural reasoning, and dual-path multimodal enhancement to improve link prediction under data-scarce settings. Extensive experiments demonstrate that LLM-derived priors provide useful relational context, while relation-aware process injection and LMF-Max post fusion further enhance entity representations. Future work will explore higher-quality MMKG construction and more domain-adaptive multimodal LLM selection and prompting strategies.

%% file: sections/appendix.tex
\appendix
\section{Training Objective}
\label{train-obj}

The model is trained using the binary cross-entropy loss:
\begin{equation}
    \begin{aligned}
    \mathcal{L}
    = - \frac{1}{|\mathcal{S}_t|}
    \sum_{(q,r,e) \in \mathcal{S}_t}
    \Big[
    & y_{e} \log(\sigma(s(e))) \\
    & + (1 - y_{e}) \log(1 - \sigma(s(e)))
    \Big],
    \end{aligned}
\end{equation}
where $\mathcal{S}_t$ denotes the training set, $y_e \in \{0,1\}$ indicates whether entity $e$ is the ground-truth tail entity, $s(e)$ is its prediction score, and $\sigma(\cdot)$ denotes the sigmoid function.

\section{Loss-Guided Dynamic Graph Sampling}
\label{sec:dynamic_sampling}

DuPLeR is jointly pre-trained on multiple MMKGs with different scales and learning difficulties. Uniform graph sampling may allocate insufficient training effort to difficult graphs, whereas sampling directly according to instantaneous losses may produce unstable distributions or overemphasize a small subset of graphs. We therefore adopt a loss-guided dynamic sampling strategy that adaptively allocates training steps while preserving a minimum sampling probability for every graph.

For each pre-training graph \(\mathcal{G}_i\), we maintain an exponential moving average (EMA) of its training loss. After a training step produces a loss \(\ell_i\) on \(\mathcal{G}_i\), the corresponding estimate is updated as
\begin{equation}
\bar{\mathcal{L}}_i
\leftarrow
\alpha \bar{\mathcal{L}}_i
+
(1-\alpha)\ell_i,
\end{equation}
where \(\alpha \in [0,1)\) denotes the EMA smoothing coefficient. Consequently, graphs that remain more difficult to optimize receive higher estimated losses.
The preliminary graph sampling probability is obtained by applying a temperature-scaled softmax to the EMA loss estimates:
\begin{equation}
\tilde{p}_i
=
\frac{
\exp\left(\bar{\mathcal{L}}*i / \tau\right)
}{
\sum*{j=1}^{G}
\exp\left(\bar{\mathcal{L}}_j / \tau\right)
},
\end{equation}
where \(G\) is the number of pre-training graphs and \(\tau>0\) controls the concentration of the sampling distribution. A smaller \(\tau\) assigns a larger proportion of training steps to graphs with higher estimated losses.
To prevent low-loss graphs from being under-sampled, we impose a minimum sampling probability:
\begin{equation}
p_i
=
\left(1-Gp_{\min}\right)\tilde{p}*i
+
p*{\min},
\end{equation}
where \(p_{\min}\) denotes the minimum sampling probability assigned to each graph. This formulation guarantees that \(p_i \geq p_{\min}\) and \(\sum_{i=1}^{G} p_i = 1\), provided that \(p_{\min} \leq 1/G\).
During the initial training steps, the EMA estimates may not yet provide a reliable measure of graph difficulty. We therefore adopt uniform graph sampling during a warm-up phase and activate the loss-guided sampling distribution once the estimates become sufficiently stable.

We pre-train DuPLeR on three MMKGs and set \(\tau=0.6\), \(\alpha=0.8\), and \(p_{\min}=0.2\) in all experiments.

\section{Effect-Size Analysis}
\label{wilcoxon}

We conduct exact one-sided paired Wilcoxon signed-rank tests using results obtained with the same three random seeds to assess whether DuPLeR consistently improves over ProLINK. Because only three paired runs are available, the tests have limited statistical power; with three nonzero paired differences, the smallest attainable exact one-sided \(p\)-value is 0.125. We therefore report the tests together with descriptive effect-size estimates.
We report additional multi-seed results for two 1-shot settings, OpenBG-IMG-R V2 and V4, as shown in Table~\ref{tab:appendix_seed_results}.

\begin{table}[htbp]
\centering

\small
\setlength{\tabcolsep}{3pt}
\begin{tabular}{@{}llcccc@{}}
\toprule
Dataset & Model & Run 1 & Run 2 & Run 3 & Mean $\pm$ SD \\
\midrule
\multirow{2}{*}{V2}
 & ProLINK & 12.4 & 11.4 & 10.6 & 11.47 $\pm$ 0.90 \\
 & DuPLeR  & 18.1 & 17.6 & 19.0 & 18.23 $\pm$ 0.71 \\
\midrule
\multirow{2}{*}{V4}
 & ProLINK & 23.5 & 22.0 & 21.6 & 22.37 $\pm$ 1.00 \\
 & DuPLeR  & 26.5 & 25.5 & 25.2 & 25.73 $\pm$ 0.68 \\
\bottomrule
\end{tabular}

\caption{Per-seed Hits@10 results (\%) of DuPLeR and ProLINK with Qwen3-8B on OpenBG-IMG-R under the 1-shot setting.
Mean and sample standard deviation are computed over three random seeds.}
\label{tab:appendix_seed_results}
\end{table}

Across OpenBG-IMG-R V2 and V4 under the 1-shot setting, DuPLeR yields absolute improvements of 6.77 and 3.37 Hits@10 percentage points over ProLINK, respectively. The exact one-sided paired Wilcoxon signed-rank tests yield \(W^{+}=6.0\) and \(p=0.125\) in both settings and therefore do not establish statistical significance at \(\alpha=0.05\). Nevertheless, all three paired differences are positive in both settings. The observed paired effect sizes are large, with Cohen's \(d_z=4.71\) for V2 and \(d_z=10.47\) for V4.

\section{Prompt Settings and Calibration Analysis}

\begin{figure}[tb]
  \centering
  \includegraphics[width=0.9\linewidth]{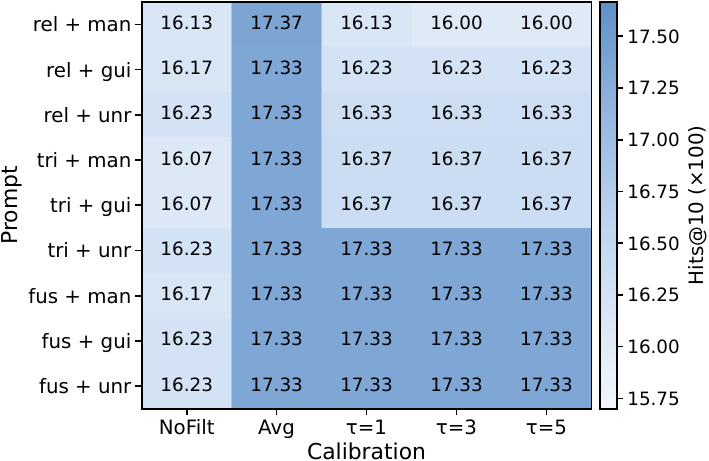}
  \caption{Prompt-configuration and calibration analysis for DuPLeR with Qwen3-VL-8B-Instruct on OpenBG-IMG-R V4 under the 1-shot setting. The prompt abbreviations follow the definitions provided in the main paper.}
  \label{heatmap_rq5}
\end{figure}

To analyze the impact of different prompt formulations and filtering thresholds on performance, we evaluated DuPLeR with the Qwen3-VL-8B-Instruct model on the OpenBG-IMG-R V4 dataset under the 1-shot setting using various combinations of prompt formulations and filtering thresholds.
Specifically, there are nine prompt configurations formed by combining three relation-information formats (Relation-only, Triple-only, and Fusion) with three entity-type constraints (Mandatory, Guided, and Unrestricted).

As shown in Figure \ref{heatmap_rq5}, the Avg calibration strategy yields consistently strong performance across all prompt configurations, with Hits@10 ranging from 17.33 to 17.37. Although Relation-only with the Mandatory constraint obtains the highest individual result of 17.37, its margin over the other configurations is only 0.04 points. More importantly, the Fusion prompt exhibits the strongest robustness under fixed-threshold filtering: all three type-constraint variants achieve 17.33 for \(\tau=1\), \(3\), and \(5\). In comparison, Relation-only prompts remain between 16.00 and 16.33, while Triple-only prompts reach 17.33 only under the Unrestricted constraint. These results suggest that combining a relation description with an example triple provides the VLM with complementary semantic and structural cues, enabling it to generate richer and more discriminative relation context.

The results also show that performance is largely insensitive to the exact value of the fixed threshold once filtering is applied, as most configurations produce identical results across \(\tau \in \{1,3,5\}\). Moreover, the differences among Mandatory, Guided, and Unrestricted constraints become negligible when the Fusion prompt is used, indicating that sufficiently informative relation context reduces sensitivity to the type-constraint formulation. By contrast, stricter constraints may limit the usefulness of the generated context when only relation descriptions or example triples are provided. Overall, the results show that both prompt composition and calibration affect downstream KGC performance when VLMs are used for relation-context enrichment. Among the evaluated configurations, Fusion provides the most stable performance across different filtering thresholds and type constraints.